\documentclass[conference]{IEEEtran}
\IEEEoverridecommandlockouts
\usepackage{cite}
\usepackage{amsmath,amssymb,amsfonts}
\usepackage{algorithmic}
\usepackage{graphicx}
\usepackage{textcomp}
\usepackage{amsthm}
\usepackage{xcolor}
\usepackage{bbding}
\usepackage{makecell}
\usepackage{algorithm}  
\usepackage{color}
\usepackage{multirow}
\usepackage{bm}
\usepackage{array}
\usepackage{booktabs}
\usepackage{tabularx,booktabs}
\usepackage{multicol}
\usepackage{cleveref}
\usepackage{balance}
\usepackage{lastpage}
\usepackage{tabulary}
\usepackage{subfigure}
\usepackage{etoolbox}
\usepackage{bbm}
\usepackage{enumitem}
\usepackage{mathrsfs}
\usepackage{multicol}
\usepackage{amsfonts,amssymb}
\usepackage{ulem}
\usepackage{cancel}
\usepackage{setspace}
\usepackage{stackengine}
\usepackage{threeparttable}

\usepackage{amsthm}
\theoremstyle{definition}

\usepackage{nomencl}
\makenomenclature

\def\BibTeX{{\rm B\kern-.05em{\sc i\kern-.025em b}\kern-.08em
    T\kern-.1667em\lower.7ex\hbox{E}\kern-.125emX}}

\begin{document}

\bstctlcite{IEEEexample:BSTcontrol}

\title{Understanding 6G through Language Models: A Case Study on LLM-aided Structured Entity Extraction in Telecom Domain
}

\author{\IEEEauthorblockN{ Ye Yuan$^1$, Haolun Wu$^1$, Hao Zhou$^1$, Xue Liu$^1$, \IEEEmembership{Fellow, IEEE}, \\
Hao Chen$^2$, Yan Xin$^2$,  Jianzhong (Charlie) Zhang$^2$, \IEEEmembership{Fellow, IEEE} }
\IEEEauthorblockA{\textit{School of Computer Science, McGill University$^1$}. 
\textit{Standards and Mobility Innovation Lab, Samsung Research America$^2$}\\
Emails:\{ye.yuan3, haolun.wu, hao.zhou4\}@mail.mcgill.ca, xueliu@cs.mcgill.ca,\\
\{hao.chen1, yan.xin, jianzhong.z\}@samsung.com} 
\vspace{-20pt} }

\maketitle

\begin{abstract}
Knowledge understanding is a foundational part of envisioned 6G networks to advance network intelligence and AI-native network architectures.  
In this paradigm, information extraction plays a pivotal role in transforming fragmented telecom knowledge into well-structured formats, empowering diverse AI models to better understand network terminologies. 
This work proposes a novel language model-based information extraction technique, aiming to extract structured entities from the telecom context. 
The proposed telecom structured entity extraction (TeleSEE) technique applies a token-efficient representation method to predict entity types and attribute keys, aiming to save the number of output tokens and improve prediction accuracy. Meanwhile, TeleSEE involves a hierarchical parallel decoding method, improving the standard encoder-decoder architecture by integrating additional prompting and decoding strategies into entity
extraction tasks.
In addition, to better evaluate the performance of the proposed technique in the telecom domain, we further designed a dataset named 6GTech, including 2390 sentences and 23747 words from more than 100 6G-related technical publications. 
Finally, the experiment shows that the proposed TeleSEE method achieves higher accuracy than other baseline techniques, and also presents 5 to 9 times higher sample processing speed. 
\end{abstract}


\begin{figure*}[t]
\centering
\setlength{\abovecaptionskip}{0pt} 
\includegraphics[width=1\linewidth]{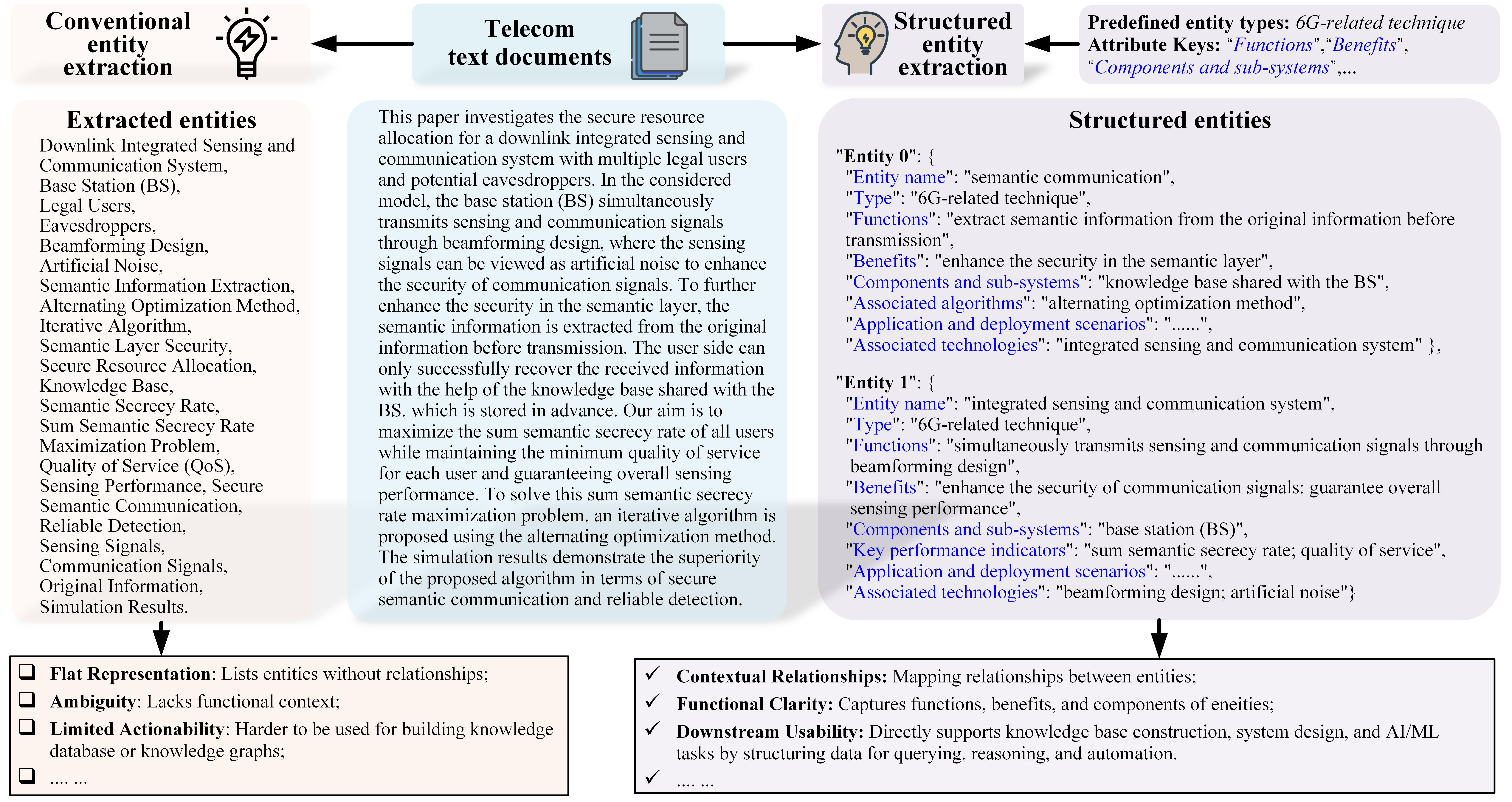}
\caption{Illustration and comparisons between conventional VS structured entity extraction in telecom context.}
\label{fig-entity}
\end{figure*}

\section{Introduction}

In the evolving 6G networks, knowledge understanding emerges as a foundational pillar to enable AI-native architectures \cite{hoydis2021toward}. Unlike previous generations, 6G demands not only high-speed connectivity but also context-aware and self-optimizing systems. This necessitates a deep semantic comprehension of telecom knowledge, such as technical specifications, network configurations, and operational protocols.

Information extraction is considered central to this paradigm to structurally organize telecom knowledge\cite{li2020survey}. 
Specifically, it aims to transform unstructured data, e.g., research papers, network logs, or equipment manuals, into machine-readable formats, such as knowledge graphs or databases. 
For example, entity extraction can parse technical documents to isolate key 6G components, mapping their dependencies within a unified knowledge base. 
With structured knowledge, entity extraction empowers AI models to understand diverse network elements in complex scenarios, e.g., network infrastructure, signal processing techniques, associated algorithms, etc.
Therefore, entity extraction serves as a cornerstone for many downstream AI and knowledge-driven tasks by identifying key elements and entities in telecom domain.

Entity extraction and recognition techniques can generally be divided into rule-based and machine learning-enabled approaches. Rule-based method usually follows specific pattern matching or dictionary lookup rules, which require predefined rules for extraction. 
Meanwhile, machine learning approaches, especially deep learning, have achieved great success in entity extraction, e.g., recurrent neural networks, long short-term memory (LSTM) networks, transformers, etc.
For instance, Bidirectional Encoder Representation from Transformers (BERT) and LSTM are integrated in \cite{wu2021construction} for entity recognition, aiming to create knowledge graphs for power communication networks. 
Similarly, BERT-LSTM schemes are used in multi-entity knowledge joint
extraction of industrial IoT equipment faults in \cite{liang2022multi}. 
BERT is also pre-trained in \cite{nimara2024entity} for telecom entity recognition, including 3GPP, customer product information, Q\&A community forum, trouble reports, etc.

The above studies have demonstrated the great potential of using machine learning methods, especially transformer schemes, for entity extraction\cite{wu2021construction,liang2022multi,nimara2024entity}. 
Different from existing studies, this work considers a more challenging task by extracting structured entities. 
In particular, structured entity extraction goes beyond identifying isolated terms, which will capture various attributes and relationships. 
Such granularity is vital for telecom systems where different entities are usually closely connected, i.e., “\textit{beamforming}” enhances “\textit{spectral efficiency}”, “\textit{edge computing}” reduces “\textit{backhaul congestion}”, using “\textit{alternating optimization}” to solve “\textit{joint active and passive beamforming}” problems. By mapping entities with their functional properties, structured extraction fuels AI-driven network automation and troubleshooting.
However, structured entity extraction is also a challenging task due to evolving network standards, sparse training data, and implicit relationships. For instance, labeling entity-attribute pairs is resource-intensive, especially for 6G domain with many emerging techniques and network elements.

The core contribution of this work is that we proposed a novel language model-enabled technique for structured entity extraction in telecom domain.
Language models have received considerable interest from the telecom industry due to their prominent capabilities in natural language processing and instruction following \cite{zhou2024large}. Existing studies have explored various applications of language models to wireless networks, including network optimization \cite{zhou2024large2}, traffic prediction \cite{hu2024self}, intent-based network management\cite{mekrache2024intent}, language models for edge intelligence, etc. 
These studies have revealed the great potential of language models for telecom tasks.

Different from existing studies \cite{wu2021construction,liang2022multi,nimara2024entity,zhou2024large,zhou2024large2,hu2024self,mekrache2024intent}, this study explores language models-enabled structured entity extraction. Specifically, the proposed telecom structured entity extraction (TeleSEE) technique first considers a token-efficient representation method, encoding entity types and attribute keys into unique special tokens. Such a representation can efficiently reduce the number of output tokens, and also improve the accuracy of entity extraction.
Meanwhile, TeleSEE applies a hierarchical parallel decoding method, introducing a multi-stage decoding strategy based on standard encoding-decoding architecture. Then each stage can be well-tuned to better adapt to the specific entity extraction process, e.g., entity names, attribute keys and values.
In addition, we also designed a structured entity dataset namely 6GTech, which provides a standard dataset to evaluate the information extraction capabilities of existing techniques in the 6G domain. 
6GTech dataset includes 2390 sentences and 23747 words from more than 100 6G-related technical publications.
It utilizes the results of multiple advanced LLMs for human verification and supplementation, guaranteeing the quality of the generated dataset.

Finally, the experiments show that the proposed TeleSEE technique achieves higher entity extraction accuracy under various evaluation metrics, and also presents a 5 to 9 times higher output efficiency than baseline approaches.

\section{Structured Entity Extraction Task Formulation for Telecom}

Given a document $d$, structured entity extraction technique aims to extract specific entities from $d$ as $\mathcal{E}=\{e_1, e_2, e_3,..., e_n \}$. 
For instance, given a telecom document such as technical reports or specifications, the entities $\mathcal{E}$ may include network devices, services, and specific signal processing techniques, i.e., user devices, base stations, antennas, URLLC services, etc.   
Then, each structured entity $e$ also includes a set of attributes $\{a_1,a_2,...,a_{|A|} \}$, in which attribute $a\in A$ and $A$ is set of all possible attributes.  
Consider a user device as an entity $e$, and the related attributes $\{a_1,a_2,...,a_{|A|} \}$ may include $a_1$ as “\textit{Supported network technologies}” and the attribute values can be “\textit{5G/LTE/VoLTE/Wi-Fi 6}”, and $a_2$ as “\textit{Connectivity status}” with possible values from “\textit{connected/idle/roaming}”, etc.
Therefore, the goal of a structured entity extraction technique is to map a document $d$ to a group of structured entities $\mathcal{E}'=\{e'_1, e'_2, e'_3,..., e'_m \}$ that is close to the ground truth results $\mathcal{E}$.

To better explain the key features of structured entity extraction, Fig. \ref{fig-entity} compares conventional entity extraction and structured entity extraction in the telecom context. We select an abstract from \cite{yang2024joint} as an example, and the left side shows the output of conventional entity extraction techniques, in which all entities are listed without relationships and in a flat representation. For instance, different kinds of entities are mixed and lack functional context, e.g., “\textit{alternating optimization method}” is listed but not tied to “\textit{secure resource allocation}” as in the document.
In addition, the limited actionability may prevent further utilization for building databases or knowledge graphs without attributes or relationships.

By contrast, the right side of Fig. \ref{fig-entity} shows 2 structured entities from the telecom text.
Given the predefined entity type “\textit{6G-related techniques}”, two entities are extracted: “\textit{semantic communication}” and “\textit{integrated sensing and communication system}”.   
Then, each entity includes multiple useful attribute keys, e.g., “\textit{Technique functions}” to illustrate the core function, “\textit{Components}” to show the possible sub-systems, “\textit{Associated technologies}” such as beamforming design and artificial noise. 
Such structured entities indicate many benefits. For example, it reveals system-level dependencies between multiple entities, e.g., sensing signals $\rightarrow$ artificial noise $\rightarrow$ security enhancement. 
Based on the defined attribute keys, it can also capture functions, benefits, and components of specific techniques, i.e., “\textit{semantic information extraction enhances security}”, which is vital for optimizing telecom systems.
Finally, such an organization can directly support many downstream tasks with structured data for querying, reasoning, and automation,  
such as knowledge base construction, system design, and machine learning-related tasks.

\section{TeleSEE-based Structured Entity Extraction for 6G Knowledge}


This section presents TeleSEE, a novel language model-aided framework designed for efficient structured entity extraction in the telecom domain. 
Our method incorporates two key innovations: \textit{(\romannumeral1)} \textit{token-efficient representation} via schema-guided special tokens, and \textit{(\romannumeral2)} \textit{hierarchical parallel decoding} that decomposes the extraction process into distinct sub-tasks.
We begin by introducing our token-efficient representation technique, followed by an in-depth explanation of the hierarchical generation process.
Subsequently, we discuss evaluation metric design and dataset construction.

\subsection{Token-Efficient Representation}
Firstly, our proposed TeleSEE method saves the output sequence length by encoding entity types and attribute keys into unique, schema-derived special tokens.
In particular, for each entity type and attribute key, we map them to a single and unique token, e.g., considering “\textit{Technique functions}” or “\textit{Associated technologies}” as single tokens to be recognized.
This representation ensures that both entity types and attribute keys can be easily predicted, significantly reducing the number of output tokens required during decoding. Meanwhile, it will also reduce the difficulty of model training and thereby improving overall efficiency.
For example, consider the original entity type ``\textit{6G-related technique}'', which may be tokenized into five tokens (``\_\textit{6}'', ``\textit{G}'', ``\textit{-}'', ``\textit{related}'', ``\_\textit{technique}'') using the T5 tokenizer. With our token-efficient representation, this phrase can be replaced with a single special token, ``\textit{ent\_type\_6G-related\_technique}''.
All special tokens are derived from a predefined schema, which is known prior to model training and fixed throughout inference.

\begin{figure}[t]
\centering
\includegraphics[width=0.95\linewidth]{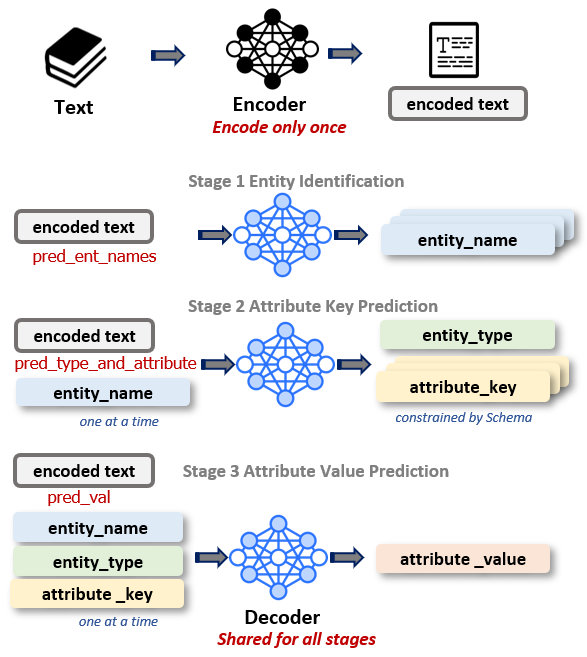}
\caption{The pipeline of our proposed method, which is built on an encoder-decoder architecture. The input text only needs to be encoded once. The decoder is shared for all the three stages. All predictions within each stage can be processed in batch. The final output of the three stages can be used to build the structured output.}
\label{fig-musee}
\end{figure}

\subsection{Hierarchical Parallel Decoding}
To further enhance efficiency and structure-awareness, TeleSEE adopts a three-stage generation pipeline.
We extend the standard encoder-decoder architecture by introducing additional prompting and decoding strategies tailored to entity extraction tasks.
In particular, given an input passage and a predefined schema, the encoder processes the input text once to generate contextualized representations. These encoded representations are then reused across all subsequent stages of the generation process. 
As illustrated in Fig.~\ref{fig-musee}, the proposed TeleSEE includes three stages:
\begin{enumerate}
    \item \textbf{Entity Identification}: In this stage, the goal is to identify entities from the text. A prompt string, \texttt{pred\_ent\_names}, is tokenized and appended to each input instance. The decoder is conditioned on this combined input to generate a sequence of entity names present in the input text.
    For instance, when focusing on 6G-related technique as shown in Fig. \ref{fig-entity}, this stage will identify the technique entity names in the text, such as “\textit{semantic communication}" and “\textit{integrated sensing and communication system}”.

    \item \textbf{Attribute Key Prediction}: For each identified entity, we predict the entity type along with its associated attribute keys. We construct a prompt of the form \texttt{pred\_type\_and\_attribute [entity\_name]}, which is tokenized, padded, and combined with the encoder output. The decoder selects tokens corresponding to predefined special tokens for types and attribute keys. This stage does not require separate classification heads—token selection is based on the probability distribution over the vocabulary. 
    Specifically, it will predict all the involved attribute keys of each entity using the token-efficient representation method, e.g., “\textit{Functions}”, “\textit{Benefits}”, and “\textit{Components and sub-systems}” in Fig. \ref{fig-entity}.  

    \item \textbf{Attribute Value Prediction}: Finally, for each $\langle$entity, attribute key$\rangle$ pair, the model predicts the associated value. Prompts are constructed in the format \texttt{pred\_val [entity\_name] [entity\_type] [attribute\_key]}. These are passed through the decoder to generate the corresponding attribute values in natural language. This step is repeated independently for each relevant $\langle$entity, attribute key$\rangle$ pair. It will predict all the attribute values of each entity, i.e., the specific “\textit{Benefits}” of entity “\textit{semantic communication}” in the given telecom context. 
\end{enumerate}

To show the idea of TeleSEE more clearly, we use Fig.~\ref{fig-musee} to illustrate the full pipeline: the input text is first encoded once, and the resulting representations are reused across all subsequent decoding stages. Each stage has a distinct objective—identifying entities, predicting their types and attribute keys, and finally generating attribute values—with stage-specific prompts guiding the decoder. The decoder architecture is shared across all stages.
During training, the decoder inputs at each stage are derived from ground truth via the teacher-forcing technique~\cite{raffel2023exploring}. The full training loss is computed as the sum of three cross-entropy losses, one for each stage.
This three-step decoding process enables modular, interpretable, and parallelizable extraction, aligning well with the structure of schema-guided entity extraction tasks.

TeleSEE is designed for both training and inference efficiency. 
Each stage is optimized to address a specific aspect of the extraction process, enabling more targeted and effective learning. 
Within a stage, predictions can be processed in parallel, significantly boosting computational efficiency. 
At inference time, segmenting long sequences into shorter spans reduces generation latency. 
Notably, TeleSEE encodes each input text only once, reusing the encoded representation across all stages. 
This design simplifies the extraction pipeline and facilitates the conversion of raw text into structured outputs.

\subsection{Evaluation Metric for TeleSEE}
\label{sec-matric}
Compared with pure entity recognition, structured entity extraction indicates higher complexity by involving multiple attributes and relationships. 
Therefore, here we introduce a new metric to justify the similarity between ground truth structured entity sets $\mathcal{E}$ and predicted results $\mathcal{E}'$:
\begin{equation}
    \delta(\mathcal{E}', \mathcal{E}) = \frac{1}{k} \bigoplus_{i,j}^{m,n} \mathbf{D}_{i,j} \cdot \Delta_{\text{ent}}(\vec{\mathcal{E}'}_i, \vec{\mathcal{E}}_j)
\end{equation}
where $m$ and $n$ are the number of predicted and ground truth entities in $\mathcal{E}'$ and $\mathcal{E}$, respectively. $k=max\{m,n\}$ indicates a average similarity metric. $\bigoplus_{i,j}^{m,n}$ is the linear sum over $\Delta_{\text{ent}}(\vec{\mathcal{E}'}_i, \vec{\mathcal{E}}_j)$, which is the pairwise entity similarities between two arbitrary entities $e\in \mathcal{E}$ and $e' \in \mathcal{E}'$.
Specifically, since $\mathcal{E}'$ and $\mathcal{E}$ may include many entities, we first implement optimal entity assignment, e.g., matching the $e\in \mathcal{E}$ and $e' \in \mathcal{E}'$ to maximize the best possible similarity. 
For instance, the entity \textit{“joint sensing and communication”} should be compared with entity \textit{“integrated sensing and communication”}, instead of the entity \textit{“semantic communication”}. 
We consider three variants for computing the optimal entity assignment:  
(i) \textbf{ExactName}: Two entities are considered a match only when their \texttt{entity name} fields are exactly equal, ignoring case sensitivity. This variant serves to evaluate performance under strict matching conditions.  
(ii) \textbf{ApproxName}: Entities are matched based on the similarity of their \texttt{entity name} fields using a soft string similarity measure. In our implementation, we adopt the Jaccard index computed over token sets. This variant enables alignment between semantically related names, such as “\textit{joint sensing and communication}” and “\textit{integrated sensing and communication}”.  
(iii) \textbf{MultiProp}: Entities are compared based on all available properties. The final similarity score is calculated as a weighted average of attribute-level similarities, where the \texttt{entity name} attribute receives a higher weight due to its greater semantic relevance. This variant is particularly useful because it can extend to scenarios where unique entity identifiers are unavailable. 
%
Then, $\mathbf{D}_{i,j}$ is 1 if entity $\vec{\mathcal{E}}_i$ and  $\vec{\mathcal{E}}_j$ are matched, and other wise $\mathbf{D}_{i,j}=0$.
Then, given the entity assignment, $\Delta_{\text{ent}}$ is defined as the linear average over individual pairwise attributes similarity $\Delta_{\text{ent}}(e', e) = \bigotimes_{p \in \mathcal{P}} \Delta_{\text{prop}}(v_{e',p}, v_{e,p})$,  where $\Delta_{\text{prop}}(v_{e',p}, v_{e,p})$ is defined as the Jaccard index between the predicted and ground-truth value tokens.


\subsection{LLM-aided 6G Entity Dataset Design}

\begin{figure}[t]
\centering
\includegraphics[width=0.85\linewidth]{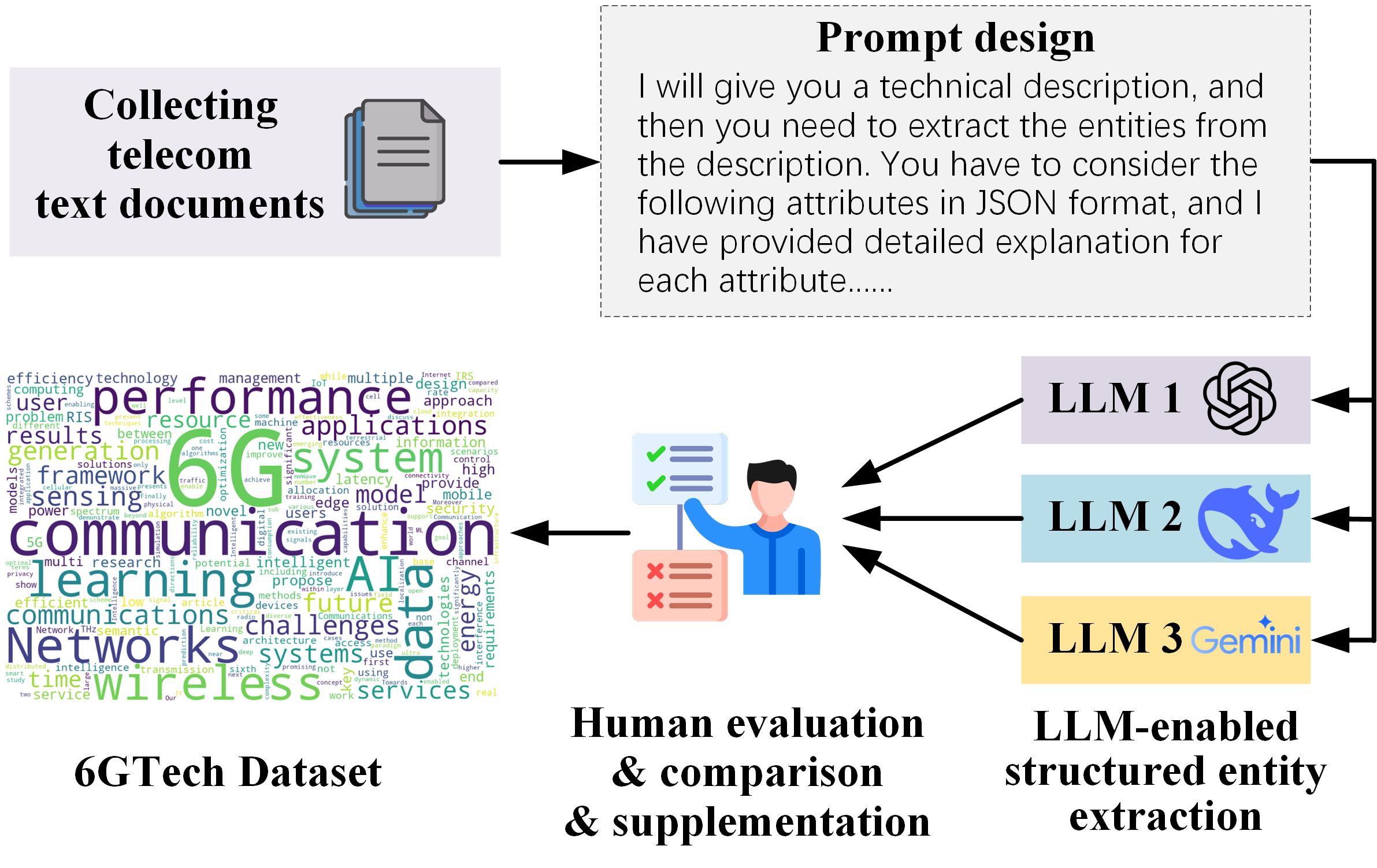}
\caption{The overall pipeline of 6GTech dataset design}
\label{fig-dataset}
\end{figure}

To better evaluate the performance of the proposed entity extraction technique, this work further designs a 6G entity dataset, namely 6GTech\footnote{The dataset is available at: https://github.com/haozhou1995/6GTech.git}. 
Fig. \ref{fig-dataset} presents the overall pipeline of the dataset design. 
Firstly, it will collect telecom text documents such as technical reports and publications. 
This work considers the abstract of technical publications as the dataset source, which are carefully selected with the keyword “6G” in the title.   
We select the publication abstract since it concisely summarizes the proposed techniques, and many useful attribute keys can be easily found as illustrated in Fig. \ref{fig-entity}, such as “\textit{Functions}”, “\textit{Components and sub-systems}”, etc.

Then, we designed an input prompt and leveraged multiple LLMs to extract structured entities from each abstract. The prompt will specify the entity types, and we select “6G-related technique” since this dataset focuses on 6G networks. It also defines all the possible attribute keys that can be extracted, ranging from “\textit{Function}”, “\textit{Components and sub-systems}”, to “\textit{Operating frequency}” and “\textit{Associated technologies}”.

Finally, we will compare the results provided by different LLMs, and further evaluate and supplement the extracted entities, e.g., missing entities, attribute keys, and misleading attribute values. 
Therefore, although LLMs are used to extract structured entities, the dataset quality is still guaranteed by human evaluation and verification.

\section{Experiments and Results}

\subsection{Experiment settings}

We compare TeleSEE against two major categories of baseline methods. 
The first group comprises models originally developed for general sequence-to-sequence (seq2seq) tasks. 
Specifically, we include (i) \textbf{LM-JSON}, where a pre-trained language model is fine-tuned to directly map a textual input into a JSON-formatted string containing the extracted entities.
The second group consists of generative models adapted from various information extraction frameworks. 
These include:
(ii) \textbf{GEN2OIE}~\cite{kolluru-etal-2022-alignment}, a two-step method that first identifies possible relations within individual sentences and then generates all corresponding entity-relation tuples;
(iii) \textbf{IMoJIE}~\cite{kolluru2020imojie}, an iterative generation model built on CopyAttention~\cite{cui2018neural}, which produces a sequence of extractions by conditioning on previously generated tuples;
(iv) \textbf{GenIE}~\cite{josifoski2022genie}, an end-to-end autoregressive framework that uses a bi-level constrained decoding mechanism to generate triplets aligned with a fixed relation schema. 
Although GenIE was originally designed for closed information extraction tasks requiring entity linking, we adapt it for our use case by disabling the entity linking component and retaining only the constrained decoding.
We adopt the pre-trained T5-Base model~\cite{raffel2023exploring} as the backbone for both our proposed TeleSEE framework and all baseline methods.
All models are trained with full parameter fine-tuning, using the Adam optimizer~\cite{kingma2017adam} with a linear warm-up schedule. The learning rate is set to $0.0001$, and the weight decay is fixed at $0.01$. Training is conducted on a single NVIDIA V$100$ GPU.

\begin{figure*}[!t]
\centering
\subfigure[Evaluation Metric with ExactName]{
\includegraphics[width=5.5cm,height=4.1cm]{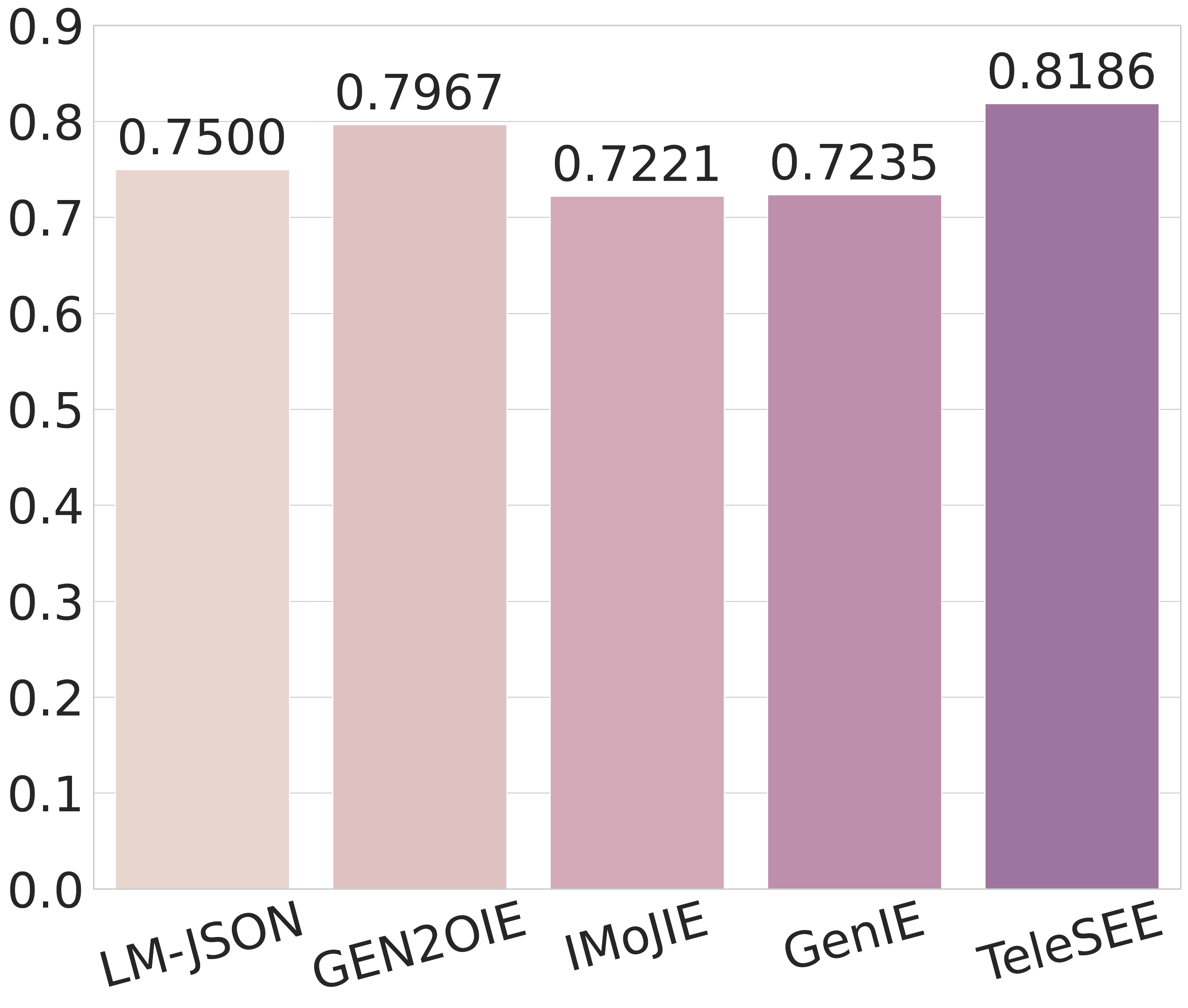} 
\label{f-r1}
}
\centering
\subfigure[Evaluation Metric with ApproxName]{
\includegraphics[width=5.5cm,height=4.1cm]{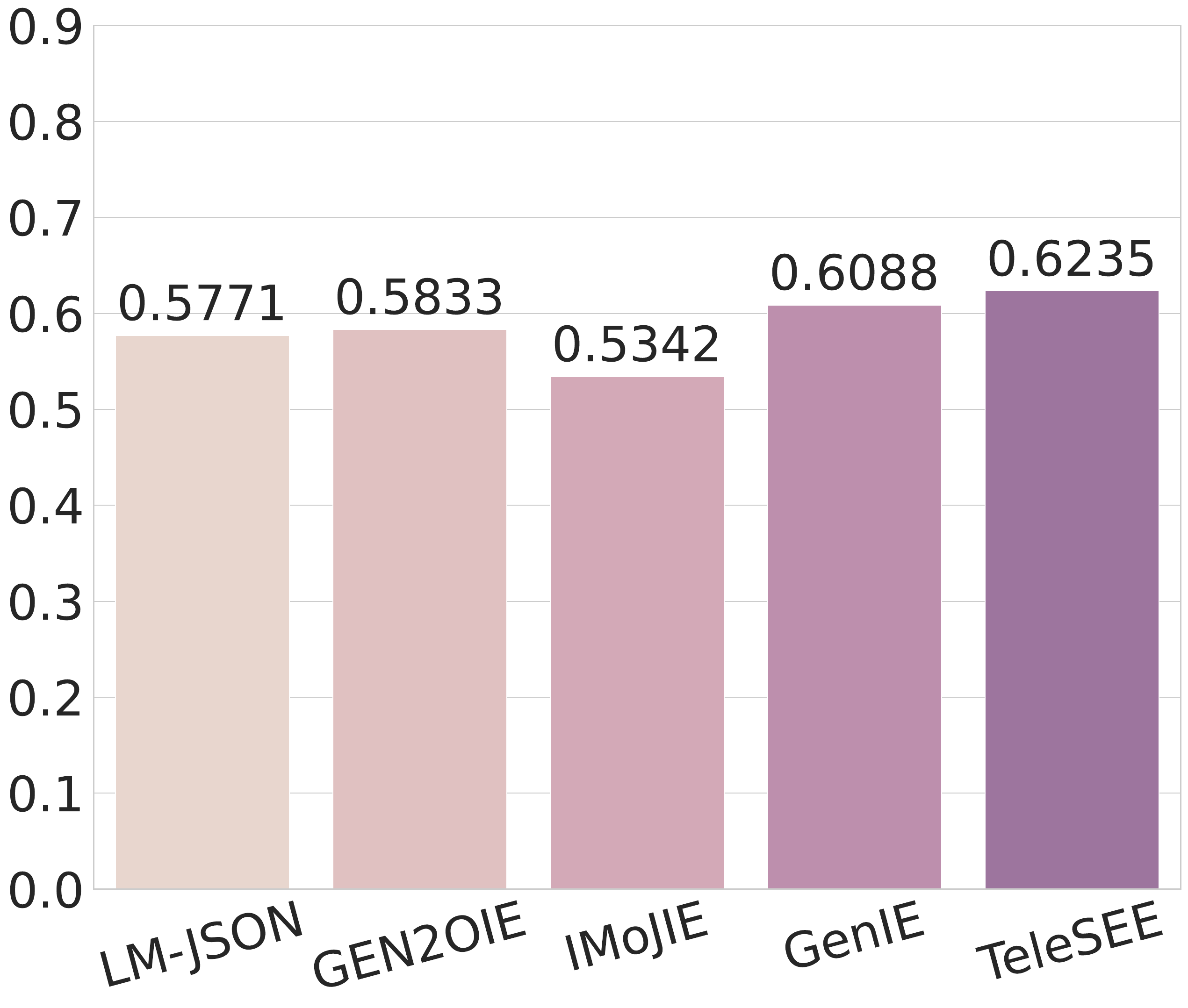} \label{f-r2}
}
\subfigure[Evaluation Metric with MultiProp]{
\includegraphics[width=5.5cm,height=4.1cm]{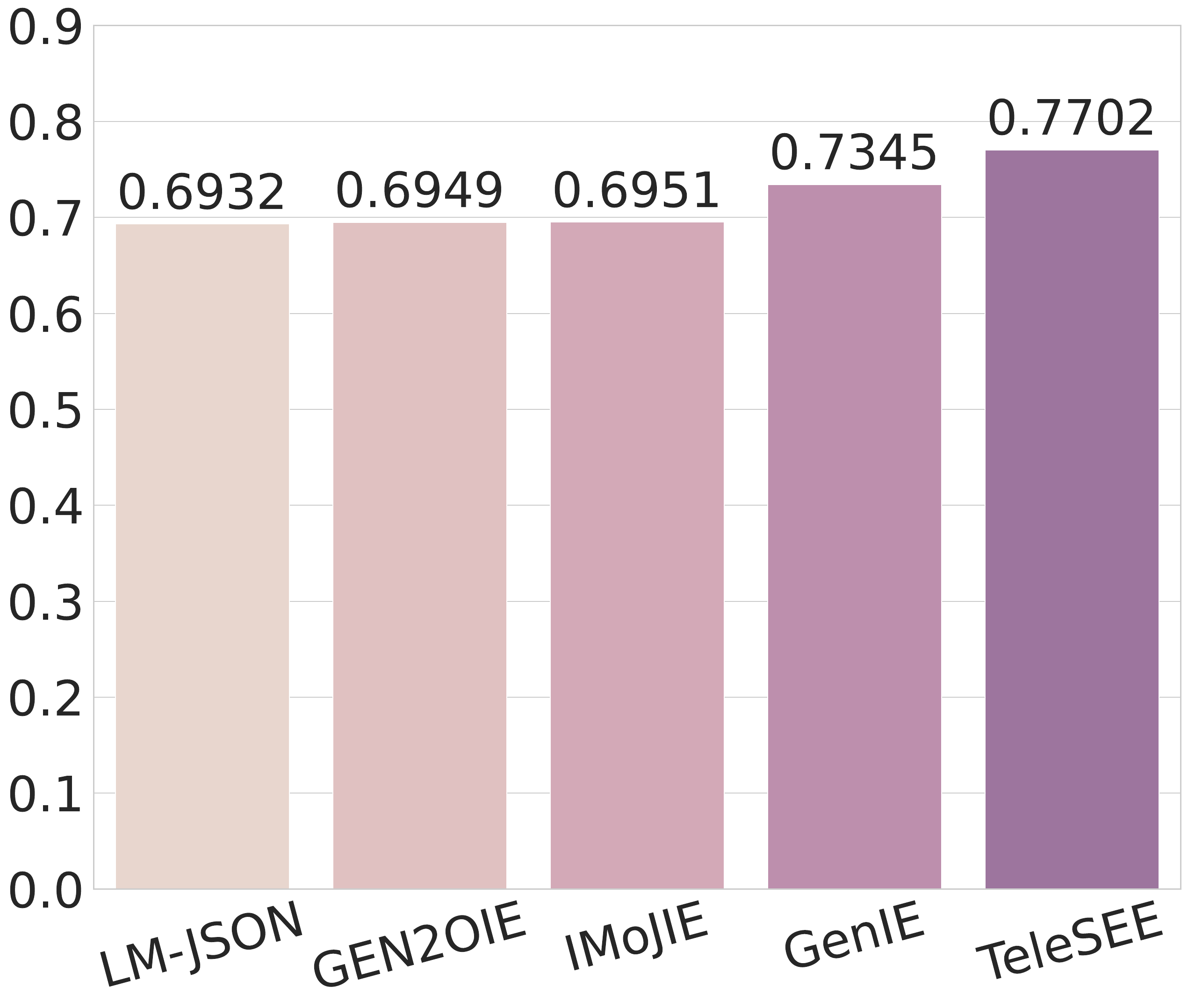} \label{f-r3}
}
\quad
\subfigure[Efficiency vs. Effectiveness]{
\includegraphics[width=5.5cm,height=4.1cm]{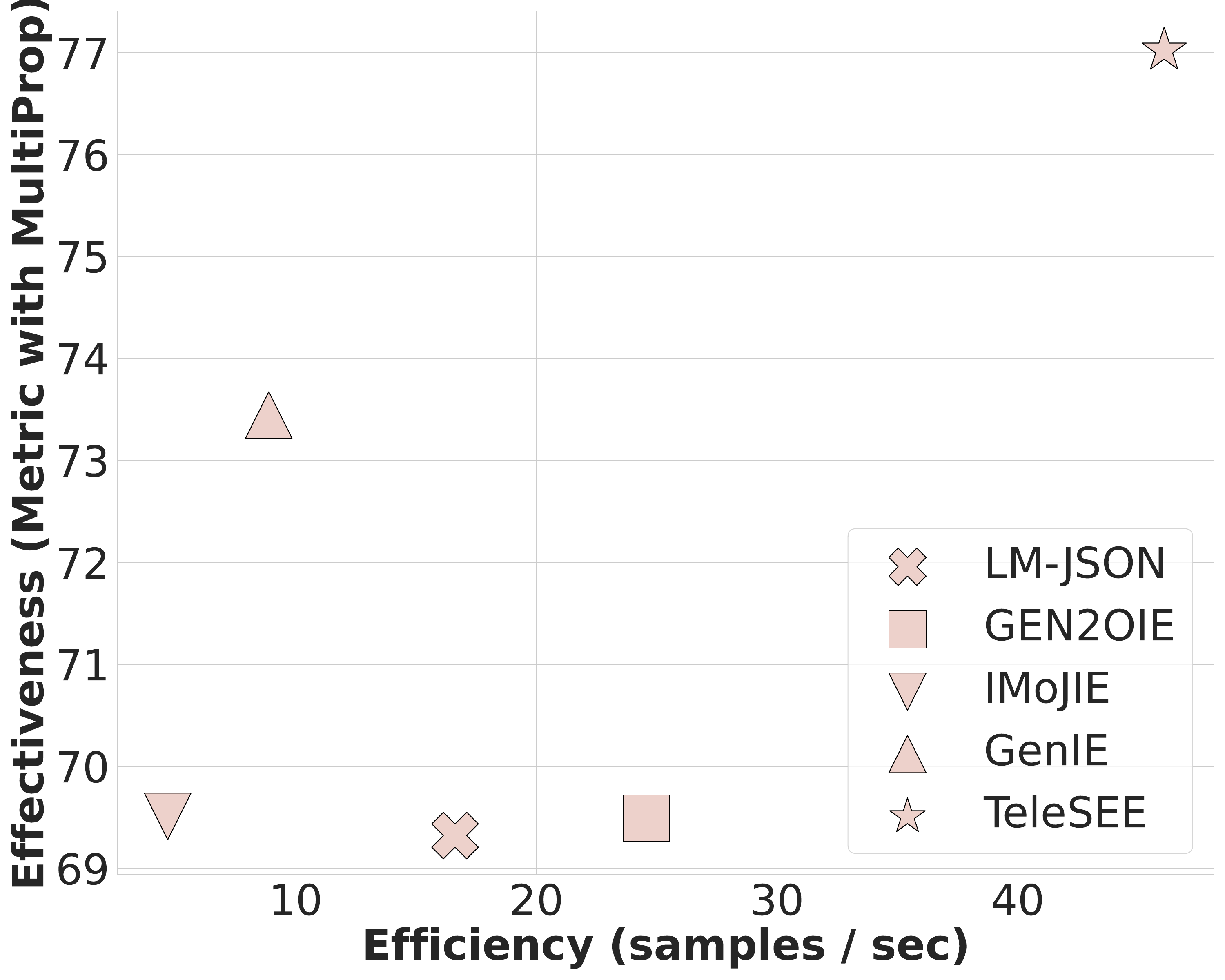} \label{f-r5}
}
\subfigure[Correlations between Three Metric Variants]{
\includegraphics[width=5.5cm,height=4.1cm]{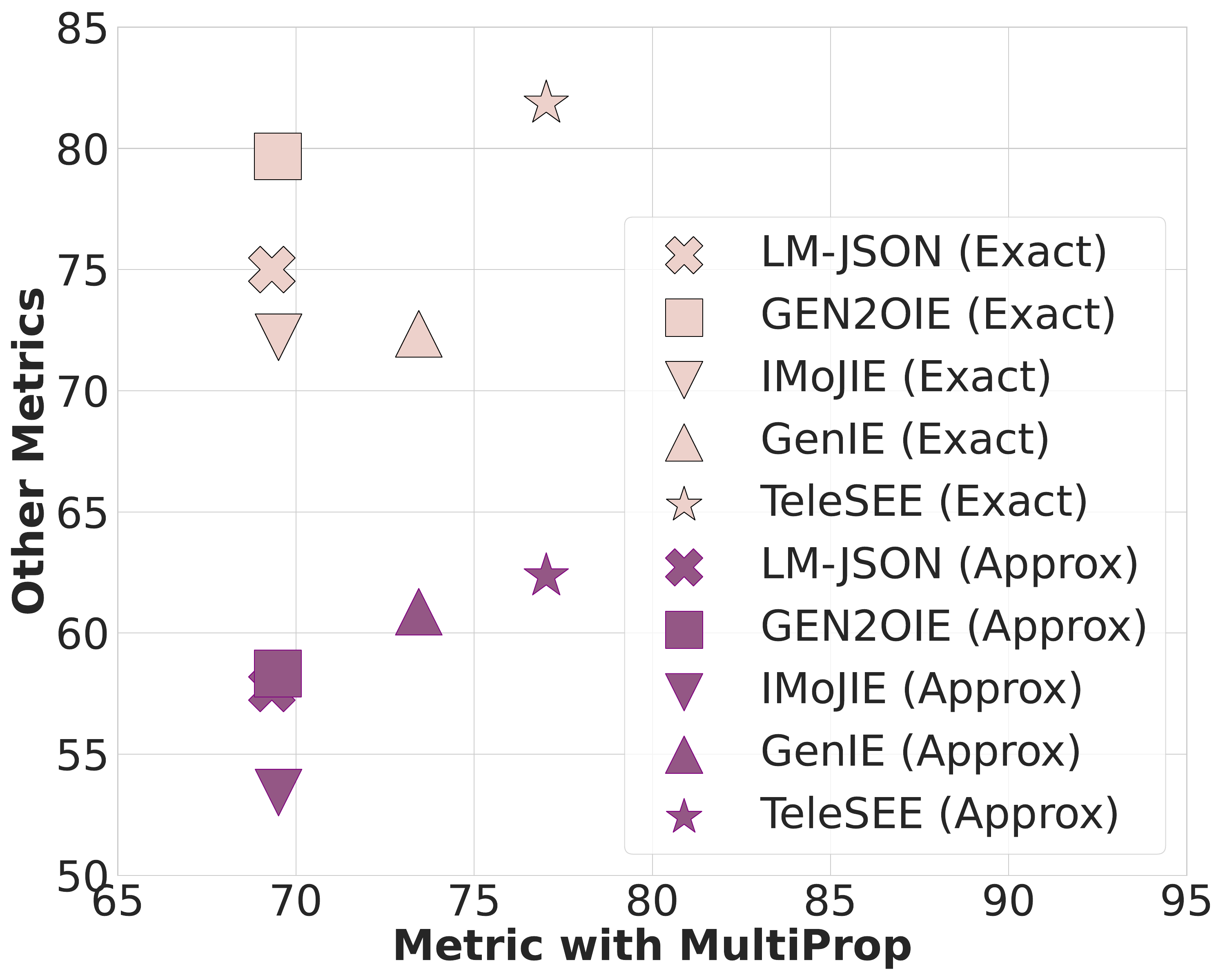} \label{f-r6}
}
\subfigure[GenIE vs. TeleSEE: Attribute-level Comparison]{
\includegraphics[width=5.5cm,height=4.1cm]{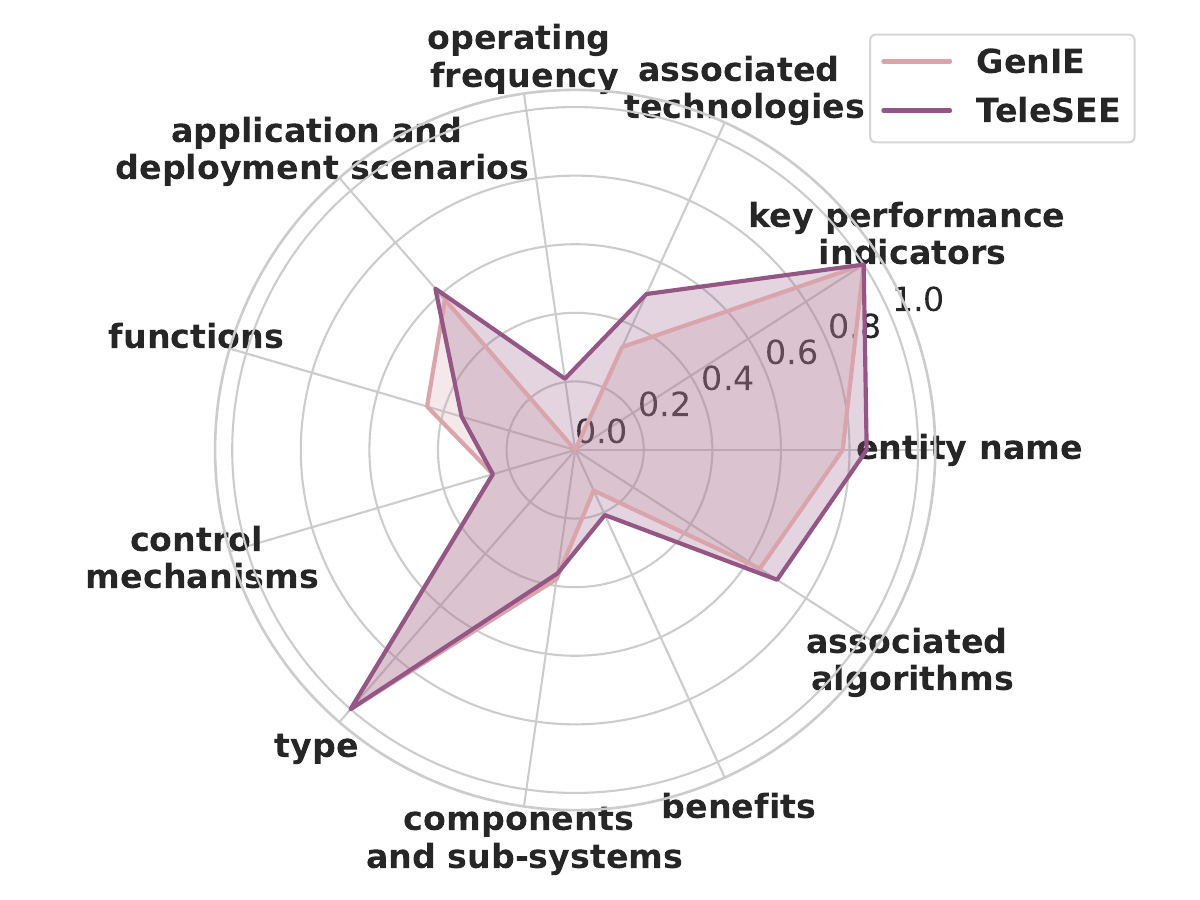} \label{f-r4}
}
\caption{Simulation results and comparisons}
\label{fig_all}
\end{figure*}

\subsection{Results}

We evaluate the performance of all baseline models and our TeleSEE method using the proposed metric under three entity matching strategies as defined in Section \ref{sec-matric}: \textit{ExactName}, \textit{ApproxName}, and \textit{MultiProp}, and the results are presented in Fig.~\ref{fig_all} (a)--(c). 
Under the strictest matching strategy ExactName, TeleSEE achieves the highest score of $0.8186$, outperforming the second best baseline GEN2OIE with $0.7967$ under ExactName.
This trend remains consistent across the other two strategies: TeleSEE achieves $0.6235$ under ApproxName and $0.7702$ under MultiProp, both the highest among all methods.
Notably, with more relaxed matching criteria, TeleSEE demonstrates robust and consistent improvements, highlighting its effectiveness in handling approximate and structure-aware cases.
We attribute this consistent performance to two core characteristics of our system. 
First, TeleSEE formulates the structured extraction task as a three-stage generation process: entity identification, attribute key prediction, and attribute value prediction.
This modular design reduces the generation search space and guides the model to attend only to relevant information at each step. 
Second, our approach leverages schema-aware prompts and special tokens, which inject inductive bias into the generation process and help the model better align its outputs with the target structure. 
In contrast, prior models such as GenIE and IMoJIE follow a monolithic generation paradigm where the entire tuple must be constructed autoregressively, often leading to noisy or misaligned outputs. 

%
In addition to accuracy, inference efficiency is critical for real-world deployment. 
Fig.~\ref{fig_all} (d) visualizes the trade-off between model efficiency, evaluated by how many samples can be processed per second, and effectiveness, quantified by the proposed metric under the MultiProp setting. 
TeleSEE achieves a processing speed of $46.08$ samples/sec, significantly faster than the strongest baseline GenIE ($8.87$ samples/sec) and more than $9$ times faster than IMoJIE ($4.67$ samples/sec). 
Moreover, this acceleration does not come at the cost of accuracy.
TeleSEE simultaneously achieves the highest MultiProp score of $0.7702$.
The efficiency benefit is primarily due to TeleSEE’s parallelizable generation pipeline. 

%
To further understand the behavior of the proposed evaluation metric under different entity matching strategies, we visualize the correlation between the three metric variants across all evaluated models in Fig.~\ref{fig_all}(e). 
We use the MultiProp variant as the reference x-axis and compare it against the ExactName and ApproxName variants on the y-axis.
We observe a positive correlation among the three variants.
While each variant targets a different matching assumption, they exhibit similar relative rankings across various methods. 
This consistency confirms the reliability of the proposed metric.

%
To provide a more fine-grained comparison, we conduct an attribute-level extracted token accuracy evaluation between TeleSEE and GenIE, the strongest baseline under both ApproxName and MultiProp variants. 
The results are presented in the radar chart in Fig.~\ref{fig_all}(f), where each axis represents one attribute type defined in the schema.
Overall, TeleSEE outperforms or matches GenIE in the majority of properties, particularly in critical fields such as \texttt{entity name}, \texttt{key performance indicators}, and \texttt{application and deployment scenarios}, demonstrating its ability to capture structured information. 
These gains highlight the strength of TeleSEE’s schema-guided, stage-wise generation mechanism, which allows it to explicitly focus on the prediction of individual attribute slots.

\section{Conclusion}
Knowledge understanding becomes increasingly important to build AI-native 6G networks, empowering diverse machine learning models and algorithms.
This work proposes a novel language model-aided information extraction method for structured entity extraction in the telecom domain. 
Additionally, it also introduces a new 6G-specific dataset, namely 6GTech, for a structured entity extraction task. 
Finally, the experiments show that the proposed TeleSEE method achieves higher accuracy and 5 to 9 times faster output efficiency than benchmarks.
In the future, we will explore more difficult and comprehensive information extraction techniques, e.g., using language models to build unified telecom knowledge bases and graphs.

\normalem
\bibliographystyle{IEEEtran}
\bibliography{Reference}

\end{document}